\documentclass[10pt,twocolumn,letterpaper]{article}

\usepackage{iccv}
\usepackage{times}
\usepackage{epsfig}
\usepackage{graphicx}
\usepackage{amsmath}
\usepackage{amssymb}

\usepackage{subcaption}
\usepackage[table,xcdraw]{xcolor}

\usepackage{booktabs}
\usepackage{rotating,tabularx}
\newcolumntype{T}{>{\raggedleft\arraybackslash} X}

\usepackage{multirow}

\usepackage{xcolor}
\definecolor{rred}{RGB}{242,72,72}
\definecolor{bred}{RGB}{181,0,0}
\definecolor{ppink}{RGB}{228,132,159}
\definecolor{yyellow}{RGB}{255,192,0}
\definecolor{bblue}{RGB}{68,114,196}
\definecolor{ppurple}{RGB}{112,48,160}
\definecolor{oorange}{RGB}{255,111,15}
\definecolor{lblue}{RGB}{157,208,237}
\definecolor{lgreen}{RGB}{179,223,136}
\definecolor{lred}{RGB}{253,152,156}

\newcommand\crule[3][black]{\textcolor{#1}{\rule{#2}{#3}}}

\usepackage{bm}

\usepackage[pagebackref=true,breaklinks=true,letterpaper=true,colorlinks,bookmarks=false]{hyperref}

\iccvfinalcopy 

\def\iccvPaperID{1342} 

\ificcvfinal\fi
\begin{document}

\title{Looking to Relations for Future Trajectory Forecast}


\author{Chiho Choi\\
Honda Research Institute USA\\
{\tt\small cchoi@honda-ri.com}
\and
Behzad Dariush\\
Honda Research Institute USA\\
{\tt\small bdariush@honda-ri.com}
}

\maketitle

\begin{abstract}

Inferring relational behavior between road users as well as road users and their surrounding physical space is an important step toward effective modeling and prediction of navigation strategies adopted by participants in road scenes. To this end, we propose a relation-aware framework for future trajectory forecast. Our system aims to infer relational information from the interactions of road users with each other and with the environment. The first module involves visual encoding of spatio-temporal features, which captures human-human and human-space interactions over time. The following module explicitly constructs pair-wise relations from spatio-temporal interactions and identifies more descriptive relations that highly influence future motion of the target road user by considering its past trajectory. The resulting relational features are used to forecast future locations of the target, in the form of heatmaps with an additional guidance of spatial dependencies and consideration of the uncertainty. Extensive evaluations on the public benchmark datasets demonstrate the robustness and efficacy of the proposed framework as observed by performances higher than the state-of-the-art methods.
\end{abstract}

\section{Introduction}
Forecasting future trajectories of moving participants in indoor and outdoor environments has profound implications for execution of safe and naturalistic navigation strategies in partially and fully automated vehicles~\cite{ammoun2009real,xu2014motion,pub3684,gao2019goal} and robotic systems~\cite{zhu1991hidden,lam2011human,kruse2013human,cao2019dynamic}.    
While autonomous navigation of robotic systems in dynamic indoor environments is an increasingly important application that can benefit from such research, the potential societal impact may be more consequential in the transportation domain.   This is particularly apparent considering the current race to deployment of automated driving and advanced driving assistance systems on public roads.  Such technologies require advanced decision making and motion planning systems that rely on estimates of the future position of road users in order to realize safe and effective mitigation and navigation strategies.

\begin{figure}
\begin{center}
 \includegraphics[width=0.47\textwidth]{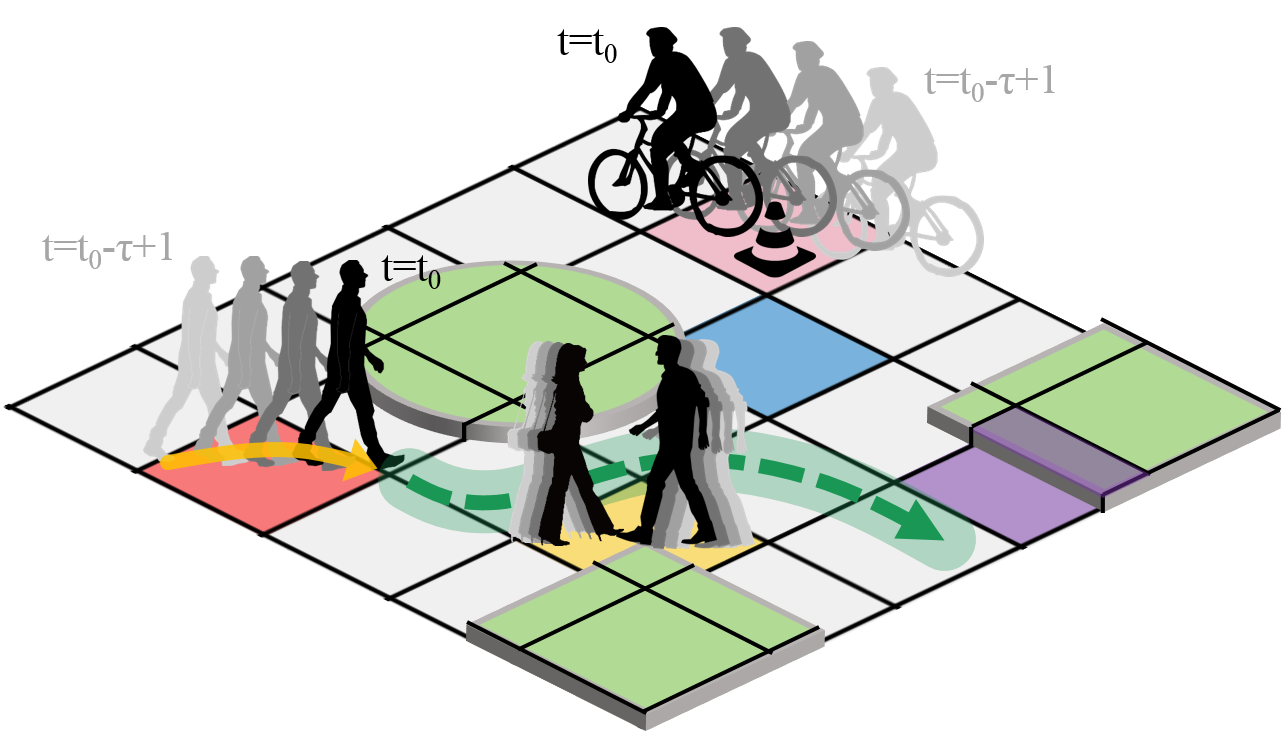}
\end{center}\vspace{-0.2cm}
   \caption{Spatio-temporal features are visually encoded from discretized grid to locally discover (i) human-human (\crule[yyellow]{0.25cm}{0.25cm}: woman$\leftrightarrow$man) and (ii) human-space interactions (\crule[rred]{0.25cm}{0.25cm}: man$\leftrightarrow$ground, \crule[pink]{0.25cm}{0.25cm}: cyclist$\leftrightarrow$cone) over time. Then, their pair-wise relations (\textit{i.e.}, \crule[rred]{0.25cm}{0.25cm}$\leftrightarrow$\crule[yyellow]{0.25cm}{0.25cm}, \crule[yyellow]{0.25cm}{0.25cm}$\leftrightarrow$\crule[ppurple]{0.25cm}{0.25cm}, \crule[ppurple]{0.25cm}{0.25cm}$\leftrightarrow$\crule[bblue]{0.25cm}{0.25cm}, \crule[bblue]{0.25cm}{0.25cm}$\leftrightarrow$\crule[pink]{0.25cm}{0.25cm}, \crule[pink]{0.25cm}{0.25cm}$\leftrightarrow$\crule[rred]{0.25cm}{0.25cm}, ...) with respect to the past motion of the target 
   (\textcolor{yyellow}{$\rightarrow$}) are investigated from a global perspective for trajectory forecast.}
\label{fig:intro}\vspace{-0.3cm}
\end{figure}

Related research~\cite{yamaguchi2011you,alahi2016social,soo2016egocentric,ma2017forecasting,su2017predicting,gupta2018social,Hasan_2018_CVPR,xu2018encoding,Yagi_2018_CVPR,sadeghian2018sophie,sadeghian2018car,yao2018egocentric} has 
attempted to predict future trajectories by focusing on social conventions, environmental factors, or pose and motion constraints. They have shown to be more effective 
when the prediction model learns to extract these features by considering human
-human (\ie, between road agents) or human-space (\ie, between a road agent and environment) interactions. 
Recent approaches~\cite{lee2017desire,xue2018ss} have incorporated both interactions to understand behavior of agents toward environments. However, they restrict human interactions to nearby surroundings and overlook the influence of distant obstacles in navigation, which is not feasible in real-world scenarios. In this view, we present a framework where such interactions are not limited to nearby road users nor surrounding medium. The proposed relation-aware approach \textit{fully} discovers human-human and human-space interactions from local scales and learns to infer relations from these interactions from global scales for future trajectory forecast.


Inferring relations of interactive entities has been researched for many years, but the focus is on the implications of relations between the object pair as in~\cite{scarselli2009graph,li2015gated}. Recently, \cite{santoro2017simple} introduced the relation network pipeline where `an object' is a visual encoding of spatial features computed using a convolutional kernel within a receptive field. Our work further expands \cite{santoro2017simple} in the sense that the word `object' incorporates spatial behavior of entities (road users, if they exist) and environmental representations (road structures or layouts) together with their temporal interactions over time, which naturally corresponds to human-human and human-space interactions (see Figure~\ref{fig:intro}). On top of this, we consider learning to infer relational behavior between objects (\textit{i.e.}, spatio-temporal interactions) for trajectory prediction.

In practice, the relations between all object pairs do not equally contribute to understanding the past and future motion of a specific road user. For example, a distant building behind a car does not have meaningful relational information with the ego-vehicle that is moving forward to forecast its future trajectory. To address the different importance of relations, the prediction model should incorporate a function to selectively weight pair-wise relations based on their potential influence to the future path of the target. 
Thus, we design an additional relation gate module (RGM) which is inspired by an internal gating process of a long-short term memory (LSTM) unit. 
Our RGM shares the same advantages of control of information flow through multiple switch gates. While producing relations from spatio-temporal interactions, we enforce the module to identify more descriptive relations that highly influence the future motion of the target by further conditioning on its past trajectory.

\begin{figure*}
\begin{center}
 \includegraphics[width=1\textwidth]{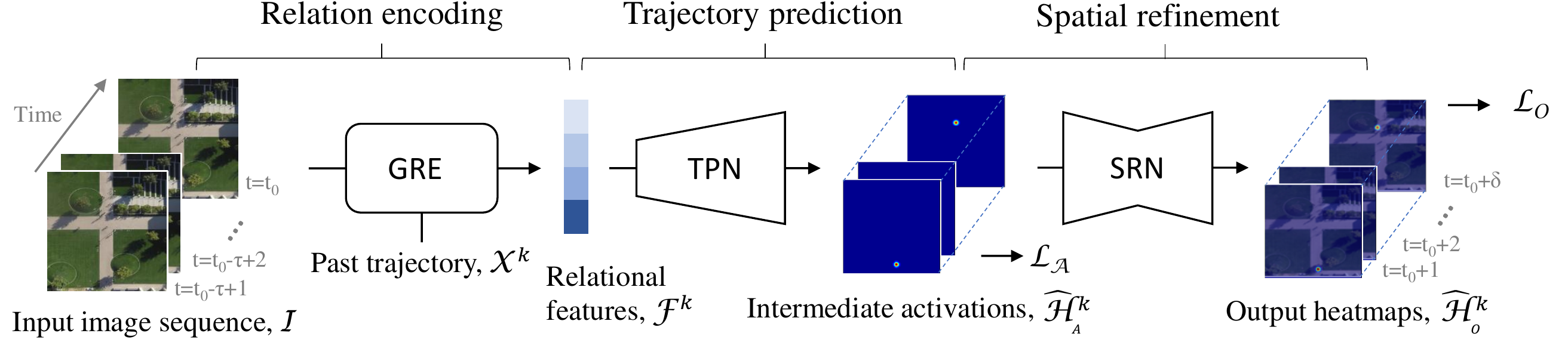}
\end{center}\vspace{-0.5cm}
   \caption{Given a sequence of images, the GRE visually analyzes spatial behavior of road users and their temporal interactions with respect to environments. The subsequent RGM of GRE infers pair-wise relations from these interactions and determines which relations are meaningful from a target agent's perspective. The aggregated relational features are used to generate initial heatmaps through the TPN. Then, the following SRN further refines these initial predictions with a guidance of their spatial dependencies. We additionally embed the uncertainty of the problem into our system at test time.
   }
\label{fig:main}\vspace{-0.3cm}
\end{figure*}

An overview of the proposed approach is presented in Figure~\ref{fig:main}. Our system visually encodes spatio-temporal features (\textit{i.e.}, objects) through the spatial behavior encoder and temporal interaction encoder using a sequence of past images (see Figure~\ref{fig:sti}). The following RGM first infers relational behavior of all object pairs and then focuses on looking at which pair-wise relations will be potentially meaningful to forecast the future motion of the target agent under its past behavior (see Figure~\ref{fig:rgm}). As a result, the gated relation encoder (GRE) produces more informative relational features from a target perspective. The next stage of our system is to forecast future trajectory of the target over the next few seconds using the aggregated relational features. Here, we predict future locations in the form of heatmaps to generate a pixel-level probability map which can be (i) further refined by considering spatial dependencies between the predicted locations and (ii) easily extended to learn the uncertainty of future forecast at test time.


The main contributions of this paper are as follows:
\begin{enumerate}
\vspace{-0.2cm}
	\item Encoding of spatio-temporal behavior of agents and their interactions toward environments, corresponding to human-human and human-space interactions.
\vspace{-0.2cm}
	\item Design of relation gating process conditioned on the past motion of the target to capture more descriptive relations with a high potential to affect its future.
\vspace{-0.2cm}
    \item Prediction of a pixel-level probability map that can be penalized with the guidance of spatial dependencies and extended to learn the uncertainty of the problem.
\vspace{-0.2cm}
	\item Improvement of model performance by $14-15\%$ over the best state-of-the-art method using the proposed framework with aforementioned contributions. 
\end{enumerate}

\section{Related Work}
\label{sec:related}

This section provides a review of deep learning based trajectory prediction. We refer the readers to~\cite{gavrila1999visual,kong2018human} for a review on recognition and prediction of human action, motion, and intention, and ~\cite{pantic2007human,rasouli2018joint} for a review on human interaction, behavior understanding, and decision making.

\noindent
\textbf{\textit{Human-human interaction} oriented approaches} Discovering social interactions between humans has been a mainstream approach to predict future trajectories~\cite{rodriguez2011data,alahi2014socially,yi2015understanding,ma2017forecasting,xu2018encoding,vemula2018social}.  Following the pioneering work~\cite{helbing1995social} on modeling human-human interactions, similar social models have been presented for the data-driven methods. A social pooling layer was proposed in~\cite{alahi2016social} in between LSTMs to share intermediate features of neighboring individuals across frames, and its performance was efficiently improved in~\cite{gupta2018social}. While successful in many cases, they may fail to provide acceptable future paths in a complex road environment without the guidance of scene context. 

\noindent
\textbf{\textit{Human-space interaction} oriented approaches} 
Modeling scene context of humans interacting with environments has been introduced as an additional modality to their social interactions. \cite{lee2017desire} modeled human-space interactions using deep learned scene features of agents' neighborhood, assuming only local surroundings of the target affect its future motion. However, such restriction of the interaction boundary is not feasible in real-world scenarios and may cause failures of the model toward far future predictions. More recently, \cite{xue2018ss} expanded local scene context through additional global scale image features. However, their global features rather implicitly provide information about road layouts than explicitly model interactive behavior of humans against road structures and obstacles. In contrast, our framework is designed to discover local human-human and human-space interactions from global scales. We locally encode spatial behavior of road users and environmental representations together with their temporal interactions over time. Then, our model infers relations from a global perspective to understand past and future behavior of the target against other agents and environments.

\noindent
\textbf{\textit{Human action} oriented approaches} These approaches rely on action cues of individuals. To predict a future trajectory of pedestrians from first-person videos, temporal changes of orientation and body pose are encoded as one of the features in~\cite{Yagi_2018_CVPR}. In parallel, \cite{Hasan_2018_CVPR} uses head pose as a proxy to build a better forecasting model. Both methods find that gaze, inferred by the body or head orientation, and the person's destination are highly correlated. However, as with human-human interaction oriented approaches, these methods may not generalize well to unseen locations as the model does not consider the road layout.


\section{Relational Inference}
We extend the definition of `object' in~\cite{santoro2017simple} to a spatio-temporal feature representation extracted from each region of the discretized grid over time. It enables us to visually discover (i) \textit{human-human interactions} where there exist multiple road users interacting with each other over time, (ii) \textit{human-space interactions} from their interactive behavior with environments, and (iii) \textit{environmental representations} by encoding structural information of the road. The pair-wise relations between objects (\textit{i.e.}, local spatio-temporal features) are inferred from a global perspective. Moreover, we design a new operation function to control information flow so that the network can extract descriptive relational features by looking at relations that have a high potential to influence the future motion of the target.

\begin{figure}
\begin{center}
\includegraphics[width=0.48\textwidth]{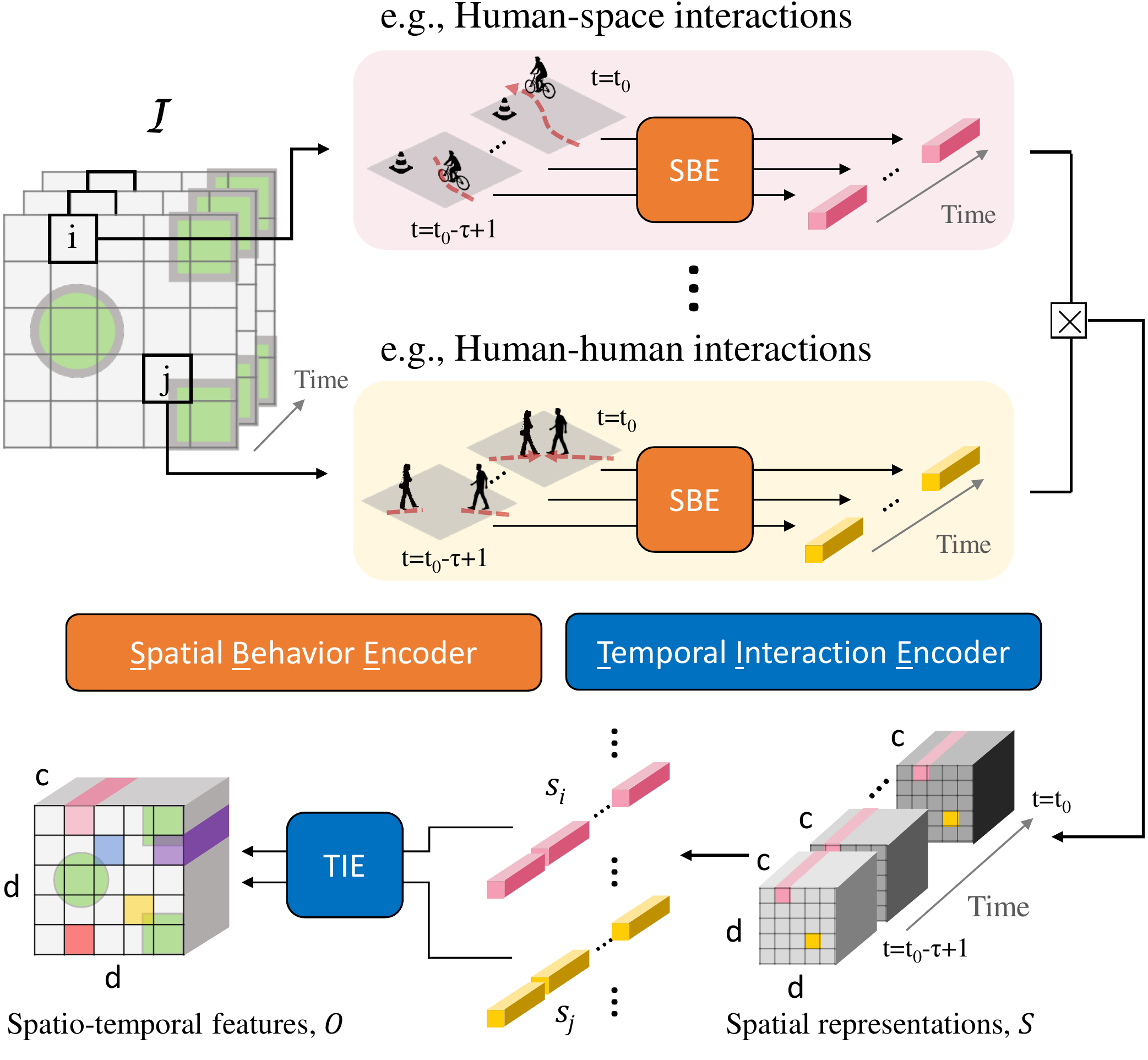}
\end{center}\vspace{-0.4cm}%
   \caption{We model human-human and human-space interactions by visually encoding spatio-temporal features from each region of the discretized grid. 
   }
\label{fig:sti}\vspace{-0.6cm}
\end{figure}

\subsection{Spatio-Temporal Interactions}
\label{sec:spatio-temporal}

Given $\tau$ past images $\mathcal{I} = \{I_{t_0-\tau+1}, I_{t_0-\tau+2}, ..., I_{t_0}\}$, we visually extract spatial representations of the static road structures, the road topology, and the appearance of road users from individual frames using the spatial behavior encoder with 2D convolutions. The concatenated features along the time axis are spatial representations $S \in \mathbb{R}^{\tau\mathsf{~x~}d\mathsf{~x~}d\mathsf{~x~}c}$. As a result, each entry $s_i \in \mathbb{R}^{\tau\mathsf{~x~}1\mathsf{~x~}1\mathsf{~x~}c}$ of $S=\{s_1,...,s_n\}$ contains frame-wise knowledge of road users and road structures in $i$-th region of the given environment. Therefore, we individually process each entry $s_i$ of $S$ using the temporal interaction encoder with a 3D convolution to model sequential changes of road users and road structures with their temporal interactions as in Figure~\ref{fig:sti}. We observed that the joint use of 2D convolutions for spatial modeling and 3D convolution for temporal modeling extracts more discriminative spatio-temporal features as compared to alternative methods such as 3D convolutions as a whole or 2D convolutions with an LSTM. Refer to Section~\ref{sec:baselines} for detailed description and empirical validation. The resulting spatio-temporal features $O\in \mathbb{R}^{d\mathsf{~x~}d\mathsf{~x~}c}$ contains a visual interpretation of spatial behavior of road users and their temporal interactions with each other and with environments. 
We decompose $O$ into a set of objects $\{o_1,...,o_n\}$, where $n=d^2$ and an object $o_i\in \mathbb{R}^{1\mathsf{~x~}1\mathsf{~x~}c}$ is a $c$-dimensional feature vector.

\begin{figure}\vspace{-0.2cm}
\begin{center}
 \includegraphics[width=0.46\textwidth]{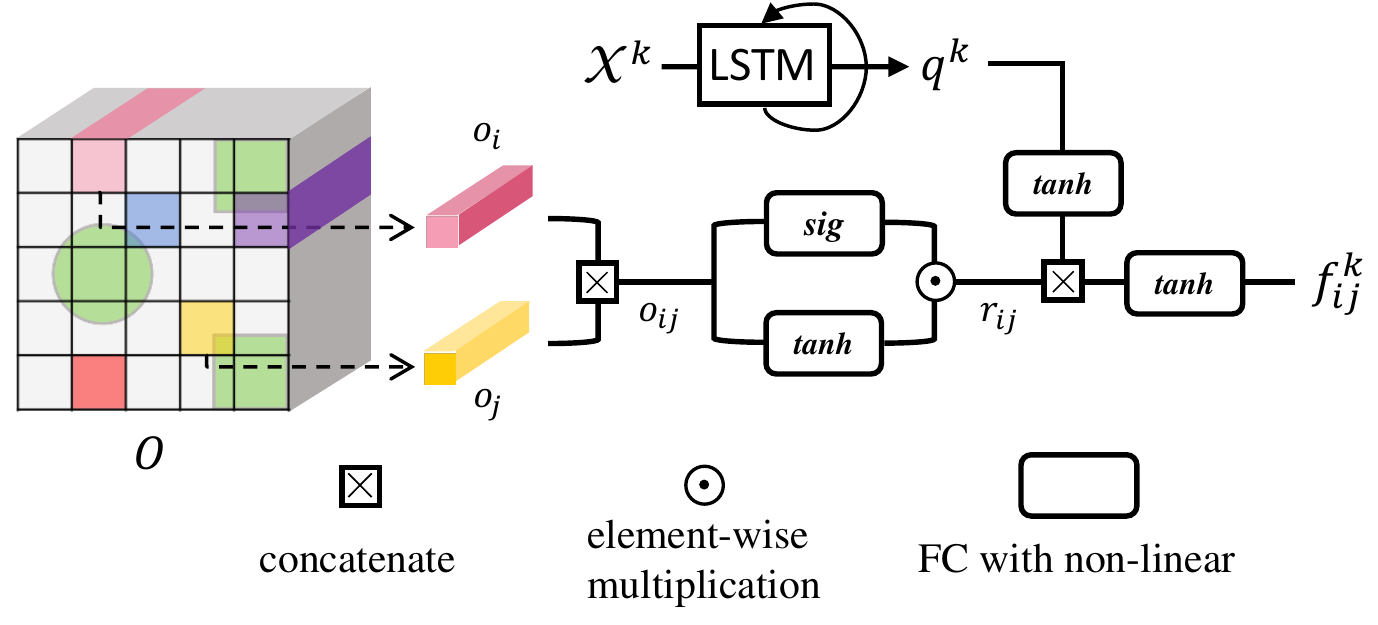}
\end{center}\vspace{-0.5cm}
   \caption{The relation gate module controls information flow through multiple switches and determines not only whether the given object pair has meaningful relations from a spatio-temporal perspective, but also how important their relations are with respect to the motion context of the target.}
\label{fig:rgm}\vspace{-0.4cm}
\end{figure}

\subsection{Relation Gate Module}
Observations from actual prediction scenarios in road scenes suggest that humans focus on only few important relations that may potentially constrain the intended path, instead of inferring every relational interactions of all road users. In this view, we propose a module which is able to address the benefits of discriminatory information process with respect to their relational importance.

We focused on the internal gating process of an LSTM unit that controls information flow through multiple switch gates. Specifically, the LSTM employs a sigmoid function with a tanh layer to determine not only which information is useful, but also how much weight should be given. The efficacy of their control process leads us to design a relation gate module (RGM) which is essential to generate more descriptive relational features from a target perspective. The structure of the proposed RGM is displayed in Figure~\ref{fig:rgm}. 

Let $g_\theta(\cdot)$ be a function which takes as input a pair of two objects $(o_i,o_j)$ and spatial context $q^k$. Note that $q^k$ is an $m$-dimensional feature representation extracted from the past trajectory $\mathcal{X}^k = \{X_{t_0-\tau+1}^k, X_{t_0-\tau+2}^k, ..., X_{t_0}^k\}$ of the $k$-th road user observed in $\mathcal{I}$. Then, the inferred relational features $\mathcal{F}^k$ are described as follows:
\begin{equation}
\mathcal{F}^k = \sum_{i,j} g_\theta (o_i,o_j,q^k),
\label{eqn:rn}
\end{equation}
where $\theta=\{\alpha,\beta,\mu,\lambda\}$ is the learnable parameters of $g(\cdot)$. Through the function $g_\theta(\cdot)$,
we first determine whether the given object pair has meaningful relations from a spatio-temporal perspective by computing 
$r_{ij} = tanh_{\alpha}(o_{ij})\odot \sigma_{\beta}(o_{ij}), $
where $o_{ij}=o_i\boxtimes o_j$ is the concatenation of two objects. Note that we add $\alpha,\beta,\mu,\lambda$ as a subscript of tanh and sigmoid function to present that these functions come after a fully connected layer. Then, we identify how their relations can affect the future motion of the target $k$ based on its past motion context $q^k$ by 
$f_{ij}^k = tanh_\lambda(r_{ij}\boxtimes tanh_\mu(q^k)). $
This step is essential in (i) determining whether the given relations $r_{ij}$ would affect the target road user's potential path and (ii) reasoning about the best possible route, given the motion history $q^k$ of the target. We subsequently collect all relational information from every pair and perform element-wise sum to produce relational features $\mathcal{F}^k\in\mathbb{R}^{1\mathsf{~x~}w}$. Note that the resulting $\mathcal{F}^k$ is target-specific, and hence individual road users generate unique relational features using the same set of objects $O$ with a distinct motion context $q^k$.

\section{Future Trajectory Prediction}

The proposed approach aims to predict $\delta$ number of future locations ${\mathcal{Y}^k} = \{{Y}_{t_0+1}^k,{Y}_{t_0+2}^k, ..., {Y}_{t_0+\delta}^k\}$ for the target road user $k$ using X$^k=\{\mathcal{I},\mathcal{X}^k\}$. Rather than regressing numerical coordinates of future locations, we generate a set of likelihood heatmaps following the success of human pose estimation in~\cite{tompson2014joint,newell2016stacked,cao2017realtime}. 
The following section details how the proposed method learns future locations.

\subsection{Trajectory Prediction Network}
\label{sec:path}

To effectively identify the pixel-level probability map, we specifically design a trajectory prediction network $a_\psi(\cdot)$ with a set of deconvolutional layers.  Details of the network architecture are described in the supplementary material.  We first reshape the relational features $\mathcal{F}^k$ extracted from GRE to be the dimension $1\mathsf{~x~}1\mathsf{~x~}w$ before running the proposed trajectory prediction network (TPN). The reshaped features are then incrementally upsampled using six deconvolutional layers, each with a subsequent ReLU activation function. As an output, the network $a_\psi(\cdot)$ predicts a set of activations in the form of heatmaps $\widehat{\mathcal{H}}_A^k \in \mathbb{R}^{W\mathsf{~x~}H\mathsf{~x~}\delta}$ through the learned parameters $\psi$. At training time, we minimize the sum of squared error between the ground-truth heatmaps $\mathcal{H}^k \in \mathbb{R}^{W\mathsf{~x~}H\mathsf{~x~}\delta}$ and the prediction $\widehat{\mathcal{H}}_A^k$, all over the 2D locations $(u,v)$. The L2 loss $\mathcal{L_A}$ is as follows: $\mathcal{L_A}= \sum_{\delta}\sum_{u,v}\left( \mathcal{H}_{(\delta)}^k(u,v)- \widehat{\mathcal{H}}_{A(\delta)}^k(u,v)\right)^2$. Note that $\mathcal{H}^k$ is generated using a Gaussian distribution with a standard deviation (1.8 in practice) on the ground-truth coordinates $\mathcal{Y}^k$ in a 2D image space. Throughout the experiments, we use heatmaps with $W=H=128$ which balances computational time, quantization error, and prediction accuracy from the proposed network structures.

\begin{figure}\vspace{-0.2cm}
\begin{center}
 \includegraphics[width=0.485\textwidth]{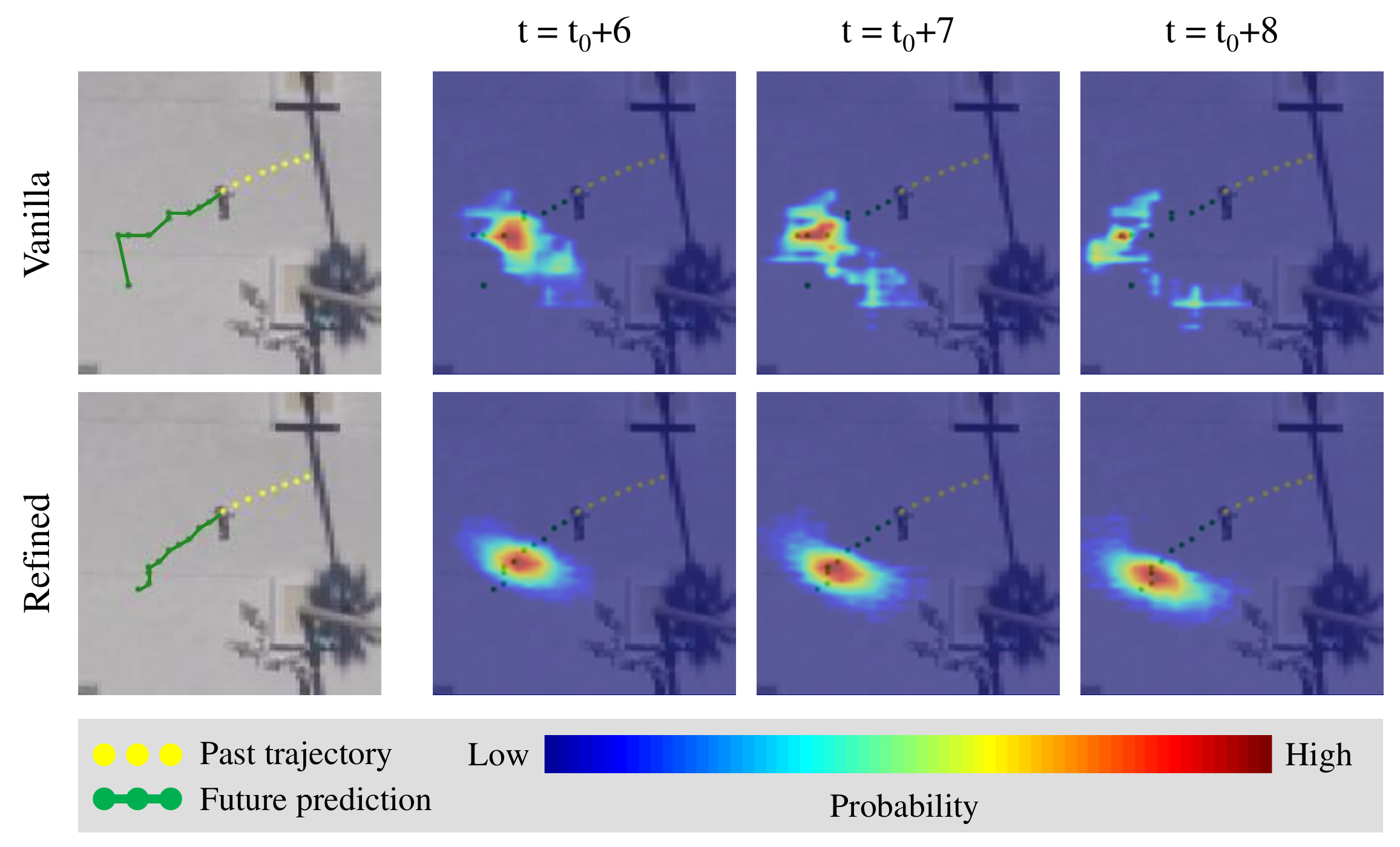}
\end{center}\vspace{-0.3cm}
   \caption{Visual analysis of spatial refinement. The first row shows the predicted future locations from the vanilla trajectory prediction network as presented in Section~\ref{sec:path}. Heatmap predictions are ambiguous, and hence the trajectory is unrealistic. The second row shows the refined locations by considering spatial dependencies as in Section~\ref{sec:refinenet}.}
\label{fig:refine}\vspace{-0.4cm}
\end{figure}

\subsection{Refinement with Spatial Dependencies}
\label{sec:refinenet}

The TPN described in the previous section is designed to output a set of heatmaps, where predicted heatmaps correspond to the future locations over time. In practice, however, the output trajectory is sometimes unacceptable for road users as shown in Figure~\ref{fig:refine}. Our main insight for the cause of this issue is a lack of \textit{spatial dependencies}~\cite{pfister2015flowing,wei2016convolutional}\footnote{Although~\cite{pfister2015flowing,wei2016convolutional} used the term for kinematic dependencies of human body joints, we believe future locations have similar spatial dependencies between adjacent locations as one follows the other.} among heatmap predictions. Since the network independently predicts $\delta$ number of pixel-level probability maps, there is no constraint to enforce heatmaps to be spatially aligned across predictions. In the literature, ~\cite{pfister2015flowing,wei2016convolutional} have shown that inflating receptive fields enables the network to learn \textit{implicit spatial dependencies} in a feature space without the use of hand designed priors or specific loss function. Similarly, we design a spatial refinement network (SRN) with large kernels, so the network can make use of rich contextual information between the
predicted locations.

We first extract intermediate activations $h_\textnormal{D5}$ from the TPN and let through a set of convolutional layers with stride 2 so that the output feature map $h_\textnormal{C17}$ to be the same size as $h_\textnormal{D2}$ (earlier activation of TPN). Then, we upsample the concatenated features $h_\textnormal{C17}\boxtimes h_\textnormal{D2}$ using four deconvolutional layers followed by a $7\mathsf{~x~}7$ and $1\mathsf{~x~}1$ convolution. By using large receptive fields and increasing the number of layers, the network is able to effectively capture dependencies~\cite{wei2016convolutional}, which results in less confusion between heatmap locations. In addition, the use of a $1\mathsf{~x~}1$ convolution enforces our refinement process to further achieve pixel-level correction in the filter space. See the supplementary material for structural details. Consequently, the output heatmaps $\widehat{\mathcal{H}}_O^k$ with spatial dependencies between heatmap locations show improvement in prediction accuracy as shown in Figure~\ref{fig:refine}.

To train our SRN together with optimizing the rest of the system, we define another L2 loss: $\mathcal{L_O}= \sum_{\delta}\sum_{u,v}\left( \mathcal{H}^k_{(\delta)}(u,v)- \widehat{\mathcal{H}}^k_{O(\delta)}(u,v)\right)^2$. Then the total loss can be drawn as follows: $\mathcal{L}_{optimize}=\zeta\mathcal{L_A}+\eta\mathcal{L_O}$. We observe that the loss weights $\zeta=\eta=1$ properly optimize our SRN with respect to the learned TPN and GRE.

\subsection{Uncertainty of Future Prediction}
\label{sec:uncertain}

Forecasting future trajectory can be formulated as an uncertainty problem since several plausible trajectories may exist with the given information. Its uncertainty has been often addressed in the literature~\cite{lee2017desire,gupta2018social,sadeghian2018sophie} by generating multiple prediction hypotheses. 
Specifically, these approaches mainly focus on building their system based on deep generative models such as variational autoencoders~\cite{lee2017desire} and generative adversarial networks~\cite{gupta2018social,sadeghian2018sophie}. As the prediction models are trained to capture the future trajectory distributions, they sample multiple trajectories from the learned data distributions with noise variations, addressing multi-modal predictions. Unlike these methods, the proposed approach is inherently deterministic and generates a single trajectory prediction. Thus, our framework technically embeds the uncertainty of future prediction by adopting Monte Carlo (MC) dropout.

Bayesian neural networks (BNNs)~\cite{denker1991transforming,mackay1992practical} are considered to tackle the uncertainty\footnote{Uncertainty can be categorized into two types~\cite{der2009aleatory}: (i) \textit{epistemic} caused by uncertainty in the network parameters and (ii) \textit{aleatoric} captured by inherent noise. We focus on epistemic uncertainty in this paper.} of the network's weight parameters. However, the difficulties in performing inference in BNNs often led to perform approximations of the parameters' posterior distribution. Recently, \cite{gal2015bayesian,gal2016dropout} found that inference in BNNs can also be approximated by sampling from the posterior distribution of the deterministic network's weight parameters using dropout. Given a dataset \textbf{X} $= \{$X$_1,...,$X$_N\}$ and labels \textbf{Y} $= \{\mathcal{Y}_1,...,\mathcal{Y}_N\}$, the posterior distribution about the network's weight parameters $\bm{\omega}$ is as follows: $p(\bm{\omega}~|~\textbf{X},\textbf{Y})$. Since it cannot be evaluated analytically, a simple distribution $q^*(\bm{\omega})$ which is tractable is instead used. In this way, the true model posterior can be approximated by minimizing the Kullback-Leibler divergence between $q^*(\bm{\omega})$ and $p(\bm{\omega}~|~\textbf{X},\textbf{Y})$, which results in performing variational inference in Bayesian modeling~\cite{gal2015bayesian}. Dropout variational inference is a practical technique~\cite{kendall2015bayesian,kendall2016modelling} to approximate variational inference using dropout at training time to update model parameters and at test time to sample from the dropout distribution $q(\bm{\omega})$. As a result, the predictive distribution with Monte Carlo integration is as follows: 
\begin{equation}\label{eqn:dropout}
\begin{split}
p(\mathcal{Y}|\textnormal{X},\textnormal{\textbf{X}},\textnormal{\textbf{Y}}) 
 & \approx \frac1{L}\sum_{l=1}^L p(\mathcal{Y}|\textnormal{X},\hat{\bm{\omega}})~~~~~~\hat{\bm{\omega}} \sim q({\bm{\omega}}),
\end{split}
\end{equation}
where L is the number of samples with dropout at test time. 

\begin{figure}
\begin{center}
 \includegraphics[width=0.46\textwidth]{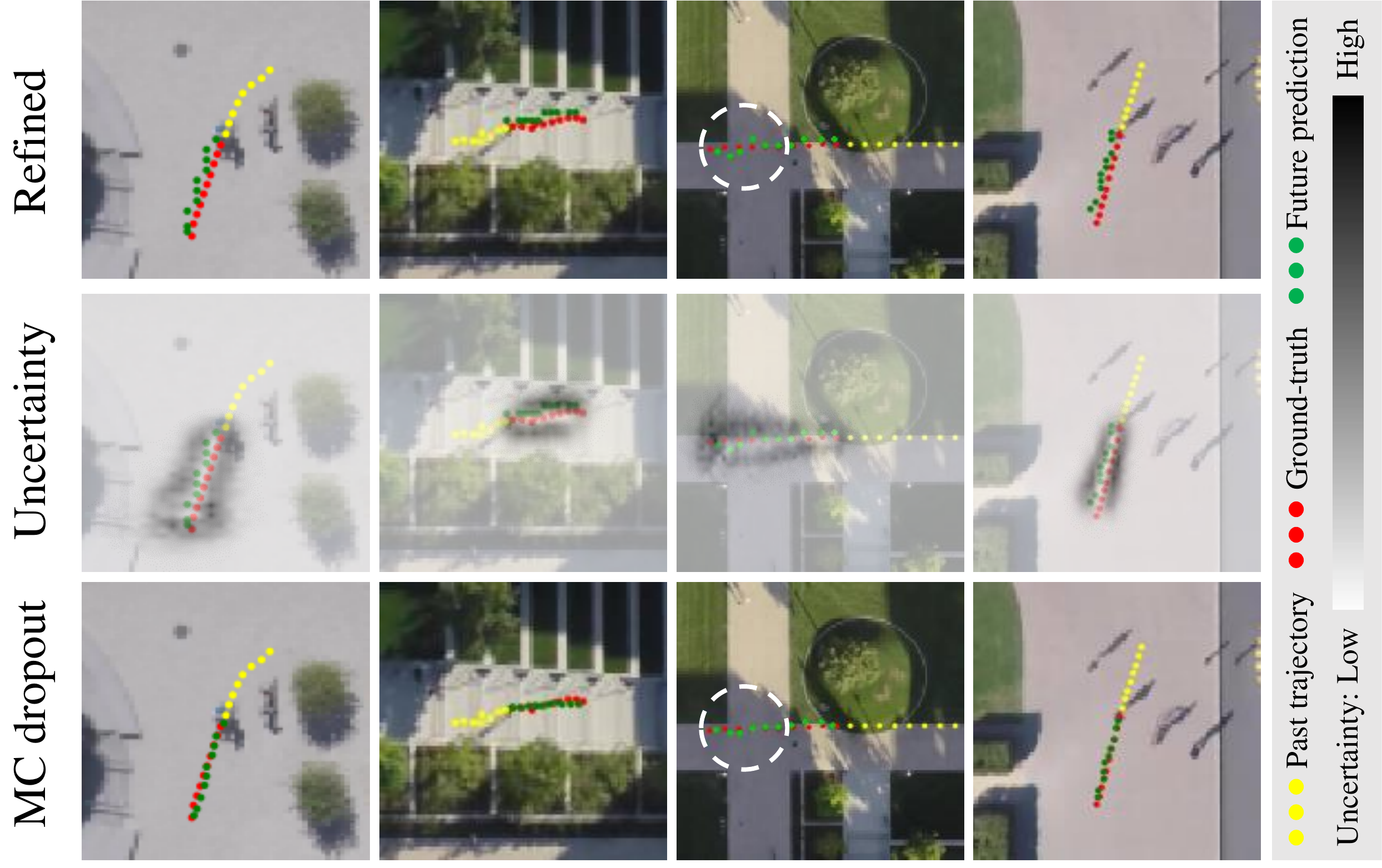}
\end{center}\vspace{-0.4cm}
   \caption{The efficacy of the uncertainty embedding into our framework. We observe that the performance of our model (first row) can be improved with MC dropout (third row). The uncertainty is visualized in the second row.}
\label{fig:uncertain}\vspace{-0.4cm}
\end{figure}

The MC sampling technique enables us to capture multiple plausible trajectories over the uncertainties of the learned weight parameters. For evaluation, however, we use the mean of $L$ samples as our prediction, which best approximates variational inference in BNNs as in Eqn.~\ref{eqn:dropout}. The efficacy of the uncertainty embedding is visualized in Figure~\ref{fig:uncertain}. We compute the variance of $L=5$ samples to measure the uncertainty (second row) and their mean to output future trajectory (third row). At training and test time, we use dropout after C6 (with drop ratio $r=0.2$) and C8 ($r=0.5$) of the spatial behavior encoder and fully connected layers ($r= 0.5$) of the RGM, which seems reasonable to balance regularization and model accuracy.

\section{Experiments}
\label{sec:exp}
We mainly use the SDD dataset~\cite{robicquet2016learning} to evaluate our approach and use ETH~\cite{pellegrini2009you} and UCY~\cite{lerner2007crowds} to additionally compare the performance with the state-of-the-art methods.

\begin{table*}
\centering
        \begin{tabular}{ll|lllll}
\hline
Category &Method & ~~~~~1.0 $sec$& ~~~~~2.0 $sec$&~~~~~3.0 $sec$&~~~~~4.0 $sec$\\
\hline 
\hline
\multirow{2}{*}{State-of-the-art} 
                              & S-LSTM~\cite{alahi2016social} & ~~~~~1.93 / 3.38&~~~~~3.24 / 5.33&~~~~~4.89 / 9.58&~~~~~6.97 / 14.57  \\
                              & DESIRE~\cite{lee2017desire} &~~~~~~~~~-~~ / \textbf{2.00}&~~~~~~~~~-~~ / 4.41&~~~~~~~~~-~~ / 7.18&~~~~~~~~~-~~ / 10.23\\
                              \hline
\multirow{4}{*}{\shortstack[l]{Spatio-temporal \\Interactions}} 
                              & \textit{RE\_Conv2D} &~~~~~2.42 / 3.09 &~~~~~3.50 / 5.23&~~~~~4.72 / 8.16&~~~~~6.19 / 11.92~~~~~ \\
                              & \textit{RE\_Conv3D}&~~~~~2.58 / 3.24 &~~~~~3.62 / 5.29&~~~~~4.83 / 8.25&~~~~~6.27 / 11.92~~~~~ \\
                              &\textit{RE\_Conv2D+LSTM} &~~~~~2.51 / 3.19 &~~~~~3.54 / 5.08&~~~~~4.60 / 7.54&~~~~~5.81 / 10.52~~~~~ \\
                              &\textit{RE\_Conv2D+Conv3D} &~~~~~2.36 / 2.99 &~~~~~3.33 / 4.80&~~~~~4.37 / 7.26&~~~~~5.58 / 10.27~~~~~ \\
                              \hline
\multirow{1}{*}{Relation Gate} 
                              & \textit{GRE\_Vanilla}&~~~~~1.85 / 2.41 &~~~~~2.77 / 4.27&~~~~~3.82 / 6.70&~~~~~5.00 / 9.58~~~~~ \\
                              \hline
\multirow{2}{*}{Spatial Refine} 
                              & \textit{GRE\_Deeper}&~~~~~2.19 / 2.84 &~~~~~3.24 / 4.88&~~~~~4.36 / 7.44&~~~~~5.63 / 10.54~~~~~ \\
                              & \textit{GRE\_Refine}&~~~~~1.71 / 2.23 &~~~~~2.57 / 3.95&~~~~~3.52 / 6.13&~~~~~4.60 / 8.79~~~~~ \\
                              \hline
\multirow{3}{*}{Uncertainty (Ours)}
                              & \textit{GRE\_MC-2}&~~~~~1.66 / 2.17 &~~~~~2.51 / 3.89&~~~~~3.46 / 6.06&~~~~~4.54 / 8.73~~~~~ \\
                              & \textit{GRE\_MC-5}&~~~~~1.61 / 2.13 &~~~~~\textbf{2.44} / 3.85&~~~~~\textbf{3.38} / 5.99&~~~~~\textbf{4.46} / 8.68~~~~~ \\ 
                              & \textit{GRE\_MC-10}&~~~~~\textbf{1.60} / 2.11 &~~~~~2.45 / \textbf{3.83}&~~~~~3.39 / \textbf{5.98}&~~~~~4.47 / \textbf{8.65}~~~~~ \\ 
                              \hline
                              
\end{tabular}\vspace{-0.2cm}
   \caption{Quantitative comparison (ADE / FDE in \textit{pixels}) of our approach with the self-generated baselines as well as state-of-the-art methods~\cite{alahi2016social,lee2017desire} using SDD~\cite{robicquet2016learning}. Note that we report our performance at 1 / 5 resolution as proposed in~\cite{lee2017desire}. 
   }
\label{tbl:baselines}\vspace{-0.2cm}
\end{table*}

\subsection{Dataset and Preprocessing}

\begin{table*}
\centering
    \begin{tabular}{l|lllll|l}
\hline
&ETH\_hotel&ETH\_eth&UCY\_univ&UCY\_zara01&UCY\_zara02&Average\\
\hline 
\hline
State-of-the-art&&&&&&\\
~~~~{S-LSTM~\cite{alahi2016social}} &0.076 / 0.125 &0.195 / 0.366&0.196 / 0.235&0.079 / 0.109&0.072 / 0.120&0.124 / 0.169 \\
~~~~{SS-LSTM~\cite{xue2018ss}} &0.070 / 0.123 &0.095 / 0.235&0.081 / 0.131&0.050 / \textbf{0.084}&0.054 / 0.091&0.070 / 0.133 \\
\hline
Ours&&&&&&\\
~~~~\textit{GRE\_Vanilla}&0.020 / 0.036 &0.054 / 0.113&0.067 / 0.129&0.050 / 0.103&0.034 / 0.067&0.047 / 0.096 \\
~~~~\textit{GRE\_Refine}&0.019 / 0.034 &0.052 / 0.100&0.065 / 0.127&0.045 / 0.086&0.031 / 0.059&0.045 / 0.086 \\
~~~~\textit{GRE\_MC-2}&\textbf{0.018} / \textbf{0.033} &\textbf{0.052} / \textbf{0.100}&\textbf{0.064} / \textbf{0.127}&\textbf{0.044} / 0.086&\textbf{0.030} / \textbf{0.059}&\textbf{0.044} / \textbf{0.086} \\
\hline
\end{tabular}\vspace{-0.2cm}
    \captionof{table}{Quantitative comparison (ADE / FDE in normalized \textit{pixels}) of the proposed approach with the state-of-the-art methods~\cite{alahi2016social,xue2018ss} using the ETH~\cite{pellegrini2009you} and UCY~\cite{lerner2007crowds} dataset. }
\label{tbl:quan2}\vspace{-0.4cm}
\end{table*}

The proposed approach aims to infer relational behavior of agents toward the environment, in addition to that against other agents. For this purpose, SDD~\cite{robicquet2016learning} fits well due to its diverse scenarios with different types of road obstacles and layouts, captured from a static platform. 
We exclude outliers following the preprocessing step in~\cite{lee2017desire}. As a result, 19.5 K instances\footnote{\cite{lee2017desire} might be more aggressively found those of unstabilized images, but we were not able to further remove outliers to match their number.} are used to train and test our model. Next, we find a center coordinate of each bounding box and use it to locate a corresponding road user onto images. Note that all RGB images are resized to fit in a 256$\mathsf{x}$256 image template, and the corresponding center coordinates are rescaled to the 128$\mathsf{x}$128 pixel space. Finally, we generate ground-truth heatmaps $\mathcal{H}$ of size 128$\mathsf{x}$128 using the rescaled center coordinates. At training and test time, we use 3.2 $sec$ of past images $\mathcal{I}$ and coordinates $\mathcal{X}^k$ of the target road user $k$ as input and predict 4.0 $sec$ of future frames as heatmaps $\widehat{\mathcal{H}}^k$. For evaluation, we first find a coordinate of a point with a maximum likelihood from each heatmap and further process the coordinates to be the same scale as original images. Then, the distance error between the ground-truth future locations $\mathcal{Y}^k$ and our predictions $\widehat{\mathcal{Y}}^k$ is calculated. We report our performance at 1 / 5 scale as proposed in~\cite{lee2017desire}.

\subsection{Comparison to Baselines}
\label{sec:baselines}
We conduct extensive evaluations to verify our design choices. Table~\ref{tbl:baselines} quantitatively compares the self-generated baseline models by measuring average distance error (ADE) during a given time interval and final distance error (FDE) at a specific time frame in \textit{pixels}.

\noindent
\textbf{Spatio-temporal interactions:} Encoding spatio-temporal features from images is crucial to discover both human-human and human-space interactions, which makes our approach distinct from others. We first conduct ablative tests to demonstrate the rationale of using spatio-temporal representations for understanding the relational behavior of road users. For this, we compare four baselines\footnote{The baselines with a prefix \textit{RE\_} do not employ the proposed gating process but assume equal importance of relations 
similarly to~\cite{santoro2017simple}.}: (i) \textit{RE\_Conv2D} which discovers only spatial interactions from $\tau$ past images using 2D convolutions; (ii) \textit{RE\_Conv3D} which extracts both spatial and temporal interactions using a well-known technique, 3D convolutions; (iii) \textit{RE\_Conv2D+LSTM} which first extracts spatial behavior using 2D convolutions and then build temporal interactions using LSTM; and (iv) \textit{RE\_Conv2D+Conv3D} where we infer spatio-temporal interactions as discussed in Section~\ref{sec:spatio-temporal}. As shown in the second section of Table~\ref{tbl:baselines}, the performance of the \textit{RE\_Conv2D+LSTM} baseline is dramatically improved against \textit{RE\_Conv2D} by replacing the final convolutional layer with LSTM. The result indicates that discovering spatial behavior of road users and their temporal interactions is essential to learn descriptive relations. It is further enhanced by using 3D convolutions instead of LSTM, as \textit{RE\_Conv2D+Conv3D} achieves lower prediction error than does the \textit{RE\_Conv2D+LSTM} baseline. This comparison validates the rationale of our use of 2D and 3D convolutions together to model more discriminative spatio-temporal features from a given image sequence. Interestingly, the \textit{RE\_Conv3D} baseline shows similar performance to \textit{RE\_Conv2D} that is trained to extract only spatial information. For \textit{RE\_Conv3D}, we gradually decrease the depth size from $\tau$ to 1 through 3D convolutional layers for a consistent size of spatio-temporal features $O$ over all baselines. In this way, the network observes temporal information from nearby frames in the early convolutional layers. However, it might not propagate those local spatio-temporal features to the entire sequence in the late layers.

\begin{figure*}
\begin{center}
 \includegraphics[width=0.99\textwidth]{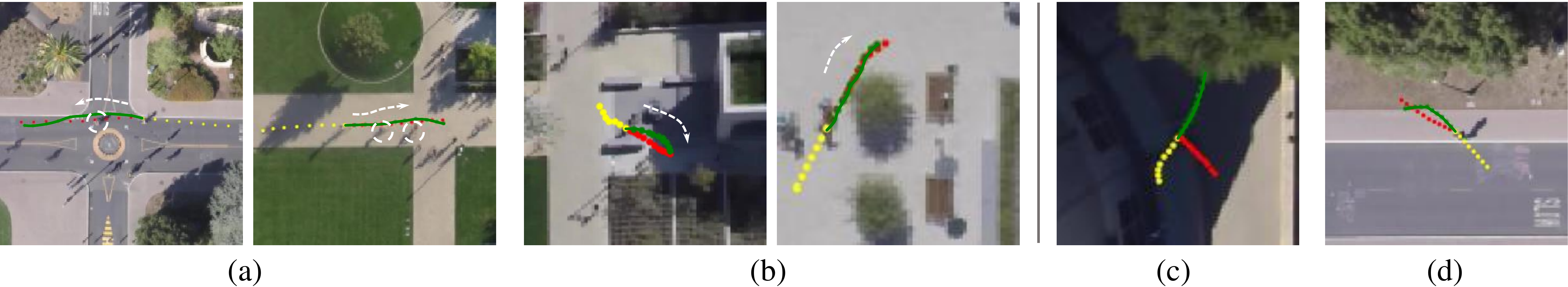}
\end{center}\vspace{-0.5cm}
\caption{The proposed approach properly encodes (a) human-human and (b) human-space interactions by inferring relational behavior from a physical environment (highlighted by a dashed arrow). However, we sometimes fail to predict a future trajectory when a road user (c) unexpectedly changes the direction of its motion or (d) does not consider the interactions with an environment. (Color codes: Yellow - given past trajectory, Red - ground-truth, and Green - our prediction)}
\label{fig:qual}\vspace{-0.1cm}
\end{figure*}

\begin{figure*}
\begin{center}
 \includegraphics[width=0.99\textwidth]{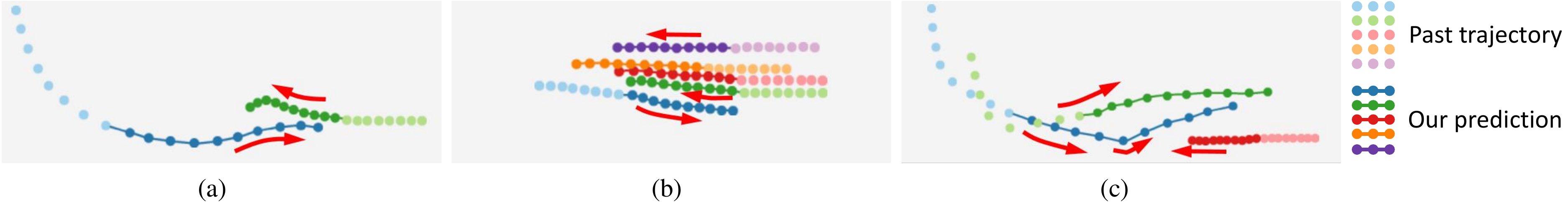}
\end{center}\vspace{-0.4cm}
\caption{Illustrations of our prediction during complicated human-human interactions. (a) A cyclist (\textcolor{lblue}{$\bullet$$\bullet$$\bullet$}) interacts with a person moving slow (\textcolor{lgreen}{$\bullet$$\bullet$$\bullet$}). (b) A person (\textcolor{lblue}{$\bullet$$\bullet$$\bullet$}) meets a group of people. (c) A cyclist (\textcolor{lblue}{$\bullet$$\bullet$$\bullet$}) first interacts with another cyclist in front (\textcolor{lgreen}{$\bullet$$\bullet$$\bullet$}) and then considers the influence of a person (\textcolor{lred}{$\bullet$$\bullet$$\bullet$}). The proposed approach socially avoids potential collisions.
}
\label{fig:qual_add}\vspace{-0.4cm}
\end{figure*}

\noindent
\textbf{Relation gate module:} To demonstrate the efficacy of the proposed RGM, we train an additional model \textit{GRE\_Vanilla} as a baseline which simply replaces the fully connected layers of \textit{RE\_Conv2D+Conv3D} with the proposed RGM pipeline. Note that we match its number of parameters to \textit{RE\_Conv2D+Conv3D} for a fair comparison. The third section of Table~\ref{tbl:baselines} validates the impact of the RGM, showing the improvements of both ADE and FDE by a huge margin in comparison to the \textit{RE\_Conv2D+Conv3D} baseline. The internal gating process of our RGM explicitly determines which objects are more likely to affect the future target motion and allows the network to focus on exploring their relations to the target road user based on the given context. The implication is that the use of the RGM is more beneficial for relational inference, and its generalization in other domains is being considered as our future work. 

\label{sec:refine}
\noindent
\textbf{Spatial refinement:} 
In addition to the qualitative evaluation in Figure~\ref{fig:refine}, we quantitatively explore how the proposed spatial refinement process helps to produce more acceptable future trajectory. The \textit{GRE\_Refine} baseline is trained using the additional spatial refinement network on top of the \textit{GRE\_Vanilla} structure. In Table~\ref{tbl:baselines}, \textit{GRE\_Refine} significantly outperforms \textit{GRE\_Vanilla} both in terms of ADE and FDE all over time. It validates that the proposed network effectively acquires rich contextual information about dependencies between future locations from initial activations $\widehat{\mathcal{H}}_A$ in a feature space. To further validate the use of the separate SRN structure, we additionally design a single end-to-end network (\textit{GRE\_Deeper}), replacing the shallow TPN of \textit{GRE\_Vanilla} with larger receptive fields and adding more layers (D1-D2 and C18-C25). Its performance  is even worse than \textit{GRE\_Vanilla}. The \textit{GRE\_Deeper} baseline experiences the difficulties in training, which can be interpreted as vanishing gradient. Thus, we conclude that the proposed approach with the separate SRN takes advantage of the intermediate supervision with two loss functions ($\mathcal{L_A}$ and $\mathcal{L_O}$), preventing the vanishing gradient problem~\cite{wei2016convolutional}.

\noindent
\textbf{Monte Carlo dropout:} 
To validate our uncertainty strategy for future trajectory forecast, we generate a set of \textit{GRE\_MC} baselines with a different suffix \textit{-L}, where \textit{L} denotes the number of samples drawn at test time. The fact that any \textit{GRE\_MC-L} baselines performs better than \textit{GRE\_Refine} certainly indicates the efficacy of the presented uncertainty embedding. By operating along with heatmap prediction, the presented approach eventually helps us to choose the points with the global maximum over the samples. Therefore, the experiments consistently show the decrease in error rate for both near and far future prediction. It is also worth noting that the use of more samples gradually increases the overall performance but introduces a bottleneck at some point as the error rate of \textit{GRE\_MC-10} is not significantly improved from \textit{GRE\_MC-5}.

\subsection{Comparison with Literature}
We quantitatively compare the performance of our models to the state-of-the-art methods using a publicly available SDD dataset~\cite{robicquet2016learning}. Two different methods are used for fair comparisons, one from \textit{human-human interaction} oriented approaches (S-LSTM~\cite{alahi2016social}) and the other from \textit{human-space interaction} oriented approaches (DESIRE\footnote{We use \textit{DESIRE-SI-IT0 Best} which shows the best performance among those without using the oracle error metric.}~\cite{lee2017desire}). In Table~\ref{tbl:baselines}, both ADE and FDE are examined from four different time steps. The results indicate that incorporating scene context is crucial to successful predictions as our methods and \cite{lee2017desire} show a lower error rate than that of \cite{alahi2016social}. Moreover, all of our models with \textit{GRE} generally outperform~\cite{lee2017desire}, validating the robustness of the proposed spatio-temporal interactions encoding pipeline which is designed to discover the entire human-human and human-space interactions from local to global scales. Note that the effectiveness of our approach is especially pronounced toward far future predictions. As discussed in Section~\ref{sec:related}, the state-of-the-art methods including~\cite{alahi2016social,lee2017desire} restrict human interactions to nearby surroundings and overlook the influence of distant road structures, obstacles, and road users. By contrast, the proposed approach does not limit the interaction boundary but considers interactions of distant regions, which results in more accurate predictions toward the far future. Note that ADE / FDE at 4 $sec$ is 5.93 / 10.56 without interactions of distant regions (worse than 5.00 / 9.58 of \textit{GRE\_Vanilla}).

In addition to the evaluation using SDD, we perform the experiments on the ETH~\cite{pellegrini2009you} and UCY~\cite{lerner2007crowds} dataset, comparing with S-LSTM~\cite{alahi2016social} and SS-LSTM~\cite{xue2018ss}. In Table~\ref{tbl:quan2}, both ADE and FDE at 4.8 $sec$ are examined in normalized $pixels$ as proposed in~\cite{xue2018ss}. Our approach mostly improves the performance over these methods, further validating our capability of interaction modeling and relational inference.


%

\subsection{Qualitative Evaluation}
Figure~\ref{fig:qual} qualitatively evaluates how inferred relations encourage our model to generate natural motion for the target with respect to the consideration of human-human interactions (\ref{fig:qual}\textcolor{red}{a}) and human-space interactions (\ref{fig:qual}\textcolor{red}{b}). Both cases clearly show that spatio-temporal relational inferences adequately constrain our future predictions to be more realistic. We also present prediction failures in Figure~\ref{fig:qual}\textcolor{red}{c} where the road user suddenly changes course and \ref{fig:qual}\textcolor{red}{d} where the road user is aggressive to interactions with an environment. Extension to incorporate such human behavior is our next plan. In Figure~\ref{fig:qual_add}, we specifically illustrate more complicated human-human interaction scenarios. As validated in these examples, the proposed approach visually infers relational interactions based on the potential influence of others toward the future motion of the target.

\vspace{-0.1cm}
\section{Conclusion}

We proposed a relation-aware framework which aims to forecast future trajectory of road users. Inspired by the human capability of inferring relational behavior from a physical environment, we introduced a system to discover both human-human and human-space interactions. The proposed approach first investigates spatial behavior of road users and structural representations together with their temporal interactions. Given spatio-temporal interactions extracted from a sequence of past images, we identified pair-wise relations that have a high potential to influence the future motion of the target based on its past trajectory. To generate a future trajectory, we predicted a set of pixel-level probability maps  
and find the maximum likelihood. We further refined the results by considering spatial dependencies between initial predictions as well as the nature of uncertainty in future forecast. Evaluations show that the proposed framework is powerful as it achieves state-of-the-art performance.

{\small
\bibliographystyle{ieee_fullname}
\bibliography{egbib}
}

\end{document}